# Critical Analysis: Bat Algorithm based Investigation and Application on Several Domains

Shahla U. Umar[1,2,*], Tarik A. Rashid[3]
[1]Technical College of Informatics, Sulaimani Polytechnic University, Sulaimani, KRG, Iraq
[2]Network Department, College of Computer Science and Information Technology, Kirkuk University, Kirkuk, Iraq.
[3]Computer Science and Engineering, School of Science and Engineering, University of Kurdistan Hewler, Erbil- KRG, Iraq.

[*]corresponding author: shahla_aothman@yahoo.com

## Abstract

In recent years several swarm optimization algorithms, such as Bat Algorithm (BA) have emerged, which was proposed by Xin-She Yang in 2010. The idea of the algorithm was taken from the echolocation ability of bats.

**Purpose-** The purpose of this study is to provide the reader with a full study of the Bat Algorithm, including its limitations, the fields that the algorithm has been applied, versatile optimization problems in different domains, and all the studies that assess its performance against other meta-heuristic algorithms.

**Design/Methodology/Approach-** Bat Algorithm is given in-depth in terms of backgrounds, characteristics, limitations, it has also displayed the algorithms that hybridized with BA (K-Medoids, Back-propagation neural network, Harmony Search Algorithm, Differential Evaluation Strategies, Enhanced Particle Swarm Optimization, and Cuckoo Search Algorithm) and their theoretical results, as well as to the modifications that have been performed of the algorithm (Modified Bat Algorithm (MBA), Enhanced Bat Algorithm (EBA), Bat Algorithm with Mutation (BAM), Uninhabited Combat Aerial Vehicle-Bat algorithm with Mutation (UCAV-BAM), Nonlinear Optimization). It also provides a summary review that focuses on improved and new Bat Algorithm (Directed Artificial Bat Algorithm (DABA), Complex-valued Bat Algorithm (CBA), Principal Component Analyses- Bat Algorithm (PCA-BA), multiple strategies coupling Bat Algorithm (MixBAT), Directional Bat Algorithm (dBA)).

**Findings-** Shed light on the advantages and disadvantages of this algorithm through all the researches that dealt with the algorithm in addition to the fields and applications it has addressed in the hope that it will help scientists understand and develop it.

**Originality/value-** As far as the research community knowledge, there is no comprehensive survey study conducted on this algorithm covering all its aspects.

**Paper type** Review paper

# 1. Introduction

The inefficiency of traditional artificial intelligence methods in keeping pace with the development in the field of optimization and machine learning in information systems with large databases, in addition to the development of the industrial revolution, especially in the fields of energy and pharmaceutical industries, all of these have led to highlight the shortcomings of artificial intelligence, and thus, opened the door for the establishment of new models in various commercial and engineering applications and the emergence of so-called computational intelligence (CI), which is known as the process of creating a computational model with tools for intelligence, which can directly entering numerical unprocessed data and then processing them and respond to problems in real-time, high reliability and with least possible errors [1].

Swarm intelligence (SI) is the ability of a swarm (large numbers of living organisms) to harmonize and cooperate within a specific environment, although there is no central leadership for them. The natural behaviours of the swarm, such as honey bees, birds swarm, and ant colonies are simulated in a mathematical model. A swarm-based algorithm is one of the significant nature-inspired algorithms that used to address optimization problems. Recently, SI algorithms have been used in many areas concerning issues of optimization, classification, and image processing, in addition to their applications in a wide range of several disciplines including grid computing of job scheduling, assignment problems, and data mining [1,2].

Depending on the setting of the algorithms, the researchers divided the optimization algorithms into two classes; the deterministic algorithms that lead to the same results when using constant primitive values at the beginning of repetition, and the stochastic algorithms, which frequently produce different results each time using the same initial values. The second class also is subdivided into heuristic and meta-heuristic algorithms. A metaheuristic algorithm is a search method with a high level of methodology that guides search agents toward the most feasible region in the search space. Metaheuristic algorithms include the genetic algorithm and the evolutionary algorithm, in addition to algorithms inspired by swarm intelligence. Recently, these algorithms have proven their efficiency in solving optimization problems, especially non-linear ones, through a balance between exploration and random search [3]. The meta-heuristic algorithms were widely used to obtain as better as an optimal solution within a reasonable time and cost. Meta-heuristic algorithms' efficiency and applying them in many real-world problems due to different features: they are uncomplicated and can be easily implemented, as well as their ability to avoid trapping in the local optimal solution [4,5].

The beginning of the evolution of nature-inspired algorithms was in the early 1960s and 1970s of the last century at the Michigan University, when the researcher John Holland's cooperation with a group of his colleagues and students, introduced the Genetic Algorithm (GA). In principle, this algorithm aimed to study natural adaptation phenomena and try to simulate them in computer systems [6]. Later, in 1983 researchers S. Kirkpatrick, C. D. Gellat, and M. P. Vecchi developed an algorithm based on the process of annealing of the metal called Simulated Annealing (SA) [7]. However, work and research in this field have been progressed significantly in recent years. For example, Particle Swarm Optimization (PSO) was one of the most powerful swarm intelligence algorithms, which was presented in 1995 by J. Kenndey and R. Eberhart [8,9] and inspired by the normal behaviour of fish or bird. Individuals in this swarm form a social network among them whereas, in 1990, researcher M. Dorigo proposed through his thesis, an optimization model for an algorithm inspired by the real behaviour of ants called (Ant Colony Optimization) [10].

In previous years, many algorithms derived from the behaviour of swarms appeared. In 1997, Differential Evolution (DE) was proposed by R. Storn and K. Price, this algorithm was used to minimize the linear and non-linear test function and proved it's outperformance on the Adaptive Simulated Annealing (ASA) as well as the Breeder Genetic Algorithm (BGA) [11]. In 2005, D. Karaboga and B. Basturk studied the ability of honeybees to forage in nature, and thus presented a new algorithm named Artificial Bee Colony (ABC) [12]. Later, in 2009, the researcher Xin-She Yang presented two algorithms, one inspired by the behaviour of the firefly called Firefly Algorithm (FA) [13], they send flashes of light for marriage purposes, sometimes for attracting prey or as a warning method from enemies. And the other named Cuckoo Search (CS) inspired by the aggressive method of breeding in some types of cuckoo birds [14]. In 2010, Xin-She Yang proposed a new nature-inspired algorithm, based on the echolocation ability of bats [15]. Moreover, after three years Bing-Yu et al found an artificial plant optimization Algorithm (APOA) based on the natural growth process of plants [16]. Then, another nature-inspired metaheuristic algorithms were proposed in 2015, the Dragonfly Algorithm (DA) [17] was suggested by Mirjalili A. S. In this algorithm, the dragonfly's behaviour in hunting small insects was simulated. One year later, the same researcher developed two new swarm-based algorithms: the Whale Optimization Algorithm (WOA) [18] and the Salp Swarm Algorithm (SSA) [19]. Furthermore, to deal with continuous non-linear optimization problems, the author M. Cortés-Toro presented a Vapour Liquid Equilibrium (VLE) algorithm in 2018 [20].

In 2019, Jaza M. Abdullah, and Tarik Ahmed presented PSO based algorithm named Fitness Dependent Optimizer (FDO), in which the weights generated from the fitness function can be used to direct the populations toward the best solution, FDO, outclassed various famous and up-to-date algorithms, such as Dragonfly Algorithm, Whale Optimization Algorithm, Salp Swarm Algorithm, Particle Swarm Optimization, and Genetic Algorithm [9]. Both Ahmed S. Shamsaldin and Tarik A. Rashid in 2109 suggested a new nature-inspired algorithm called Donkey and Smuggler Optimization (DSO) inspired by the normal activity of donkeys in finding their routes. DSO tackles different problems of pathfinding, such as the travelling salesman problem (TSP), packet routing, and ambulance routing. The conclusions on the tests displayed superior management and searching ability of DSO against ACO and others [21]. In 2020, Bo Yanga and others presented a comprehensive study of 28 meta-heuristic algorithms compare their performance using photoelectric parameters and to help researchers in applying them in different applications [22].

Finally, both Chnoor M. Rahman and Tarik A. Rashid in 2020 produced a new algorithm called Leaner Performance-based Behavior (LPB). LPB is inspired by the procedure of accommodating graduated learners from high school in numerous departments in the university. LPB is assessed against standard benchmark functions, CEC-C06 2019 test functions, and a real application. Then, the outcomes of LPB are evaluated against the DA, GA, and PSO. The LPB generated greater results. LPB has an abundant capability to tackle large optimization problems comparing to the DA, GA, and PSO. The overall outcomes showed the ability of LPB in enhancing the initial population and coming together in the direction of the global optima. Furthermore, the outcomes of the LPB are substantiated statistically [23]

As mentioned in the above that Yang in [15] developed a new metaheuristic nature-inspired algorithm for various optimization problems called Bat Algorithm. This algorithm is derived from the echolocation behaviour of microbats. These animals have an advanced ability of hunting by sending echolocation signals, as well as their fantastic directing technique through which can distinguish between prey and other barriers even in complete darkness.

To the knowledge of the authors, there is no well-organızed survey on Bat Algorithm covering all its aspects that have been conducted tıll now. All published researches and studies performed on the Bat Algorithm since its submission by the researcher Yang in 2010 were presented in Figure (1), which indicates the variation in the number of publications related to this algorithm during the last ten years which reached its peak in 2015.

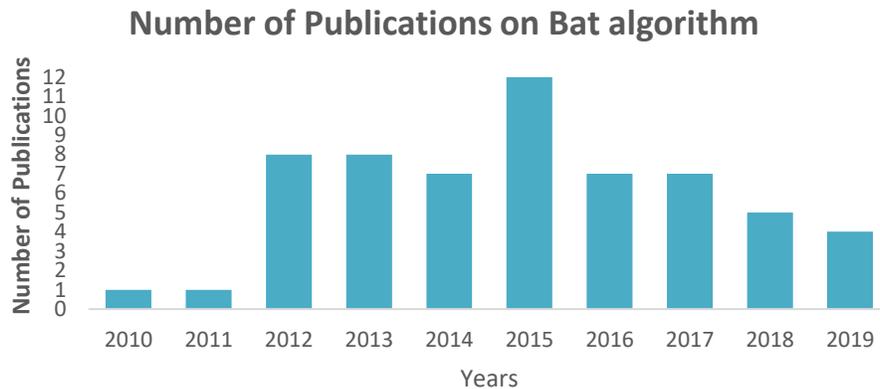

**Figure (1): The most yearly publications on the BA since 2010**

The organization of this paper starts with a general review of Bat's biological nature in hunting followed by a brief report of Bat algorithm, how it works, its main features and equations, as well as its original pseudocode. After that, most applications that were solved by the Bat Algorithm are given. In the next section, the papers that assessed the performance of the Bat Algorithm against other meta-heuristic algorithms are provided. Then, all the research works conducted on the hybridized Bat Algorithm are listed. In the next two sections, all the studies that address the modification and improvement papers on Bat Algorithm are discussed. In section 6, novel Bat Algorithms for different optimization purposes inspired by the standard Bat Algorithm is reported. Finally, the paper terminates with the conclusions of this survey and the possible research trends in the future.

## 2. Overview of Biological Inspirations of Bat

Bats form an enormous group of mammals (about one-fifth of the mammals of the world). Their great abilities of flight at night and orientation in complete darkness helped them greatly to take advantage of the environment at night and find their prey. Their search mechanism is dependent on what is called echolocation. In the late 18th century, for the first time, the scientist Lazzaro Spallanzani clearly explained the Bats' ability to fly in the night time. After 150 years, other scientists (Donald Griffi from U. S. A. and Sven Dijkgraaf from the Netherland) described the echolocation in more detail and how the Bats use it to orient in the darkness. Bats send sounds at high frequencies that exceed the upper limit of human hearing, then listen to the echo of these frequencies reflected from surrounding objects or prey. Although it is not possible to hear these ultrasound frequencies by humans, they can be analyzed and transferred to the audible range through special electronic devices [24] (Figure (2)).

All bats do not use the same echolocation signals. There are three types of these signals (Figure (3)). The first and the most popular type is the Frequency Modulated (FM), this short signal covers an expanded frequency band, which often exceeds one octave, while the second band consists of short sloped modulated followed by a constant but longer frequency band denoted by (FM-CF). Finally, the third signal is the longest fixed sound signal that usually precedes and is followed by short Frequency Modulated (FM-CF-FM). Bats have a special auditory system supported with a high-performance filter mechanism, this filter receives the tuned echo component (CF) reflected from obstacles and efficiently distinguish the audible sonic and isolate it from the background noise. Some species of Bats are very smart hunters; they can also forage under difficult conditions even in the intensive forest. So, they are forced to cope with an environment crowded with sounds and noise reflected from nature. Usually, they stand on a branch of a tree monitoring their surroundings and waiting for prey. The long, pure signal returns almost without being affected by the surrounding noise, but, when any prey flies in the echolocation frequency range, it is detected by the vibrations at the frequency generated by its wings [24].

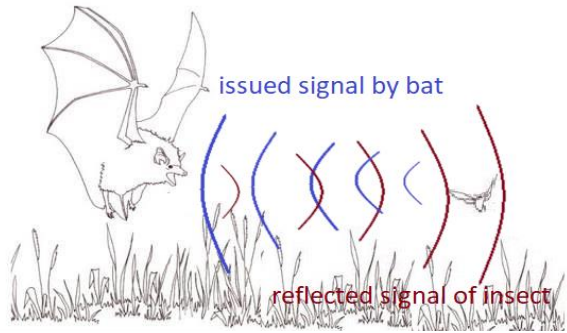

**Figure (2): Bat's echolocation [24]**

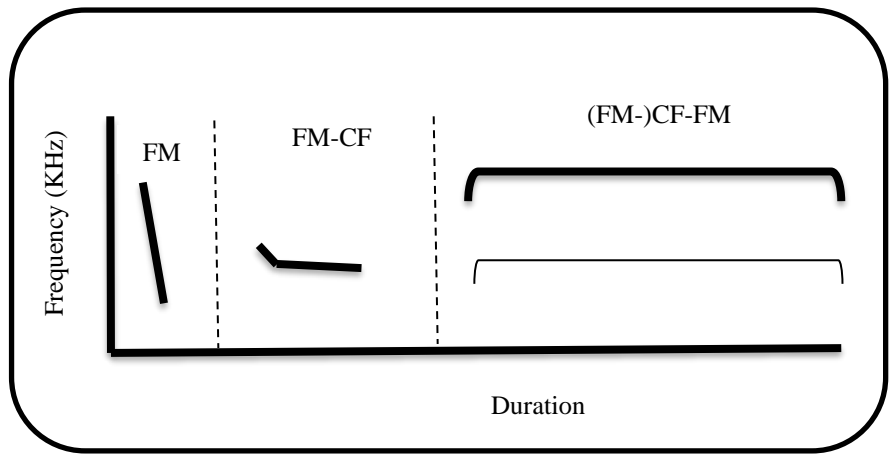

**Figure (3) Types of Bats echolocation**

## 3. The Standard Bat Algorithm

The meta-heuristic Bat Algorithm presented by Yang in 2010 took advantage of the ability of bat to find their prey by issuing echolocation signals, before clarifying more details about this algorithm, we give a brief explanation of echolocation ability [15,25]:

### 3.1. Echolocation of Microbats

Mostly Frequency Modulation is changed in their frequency and amplitude components by reflecting from the background objects, which causes a noisy echo. However, the echolocation signals generated by the bat always returned free from background noise. As well as, the bats have high hunting skills, they usually cling on a tree branch to observe the flying prey, so, once prey flying in the echolocation field of the bat, it is discovered immediately through the change in the frequency of the returned signal caused by the movement of the prey wings. The bat thus recognizes the presence of flying insects in its field. Bats can determine the location of the objects in its surroundings through the time difference between the sent and received signal pulse and also can identify its forms, types, and the direction of its movement through the relative amplitude of the sound waves captured from each ear [24].

## 3.2. Bat Algorithm

Based on the principle of echolocation behaviour mentioned above, the researcher Yang in 2010 developed the Bat Algorithm for various optimization problems. So that the algorithm is designed and used to optimize the fitness function in a way that simulates the behaviour of bats in nature.

To simulate the bat's echolocation behaviour and use it to design an algorithm, Yang in 2010 proposed several rules [15]: To estimate the distance, bats exploit their echolocation ability, also they can differentiate between insects and background obstacles. To hunt prey, bats always fly randomly with a certain velocity $(v_i)$ at a location $(x_i)$ that has a constant frequency $(f_{min})$ but variable wavelength $(\lambda)$ and loudness $(A_0)$. Bat's pulse loudness may differ in different ways. However, Yang assumed that it ranged from a positive large number $(A_0)$ to a minimum constant number $(A_{min})$.

To generate a new solution in the simulation, the frequency, loudness, and pulse rate have been tuned ((Equations (1)–(3)) and the obtained solution is regarded better than the previous one depending on the fineness of the solutions adjusted by the loudness and the pulse rate, Which are in turn depends on how close it is to the optimal solution.

$$f_i = f_{min} + (f_{max} - f_{min})\beta \qquad (1)$$

$$v_i^t = v_i^{t-1} + (x_i^t - x_*)f_i \qquad (2)$$

$$x_i^t = x_i^{t-1} + v_i^t \qquad (3)$$

From the above equations, $(x_i)$ was used to represent the bat's position in the search space, while the $(v_i)$ and $(f_i)$ used to denote the velocities and the frequency of the pulse respectively. Whereas $(\beta)$ refers to a vector of random numbers between 0 and 1. As well as, $(x_*)$ indicates the best solution obtained till now. The optimization problem's dimensions determine the upper limit and the lower limit of the frequency. At the first iteration, each bat is assigned a random number of the frequency, and the bats moved to the new positions using their new velocities. When the bats draw near their prey, the rate of $(A_i)$ reduces whereas the value of $(r_i)$ increases, as shown in the following two equations:

$$A_i^t = \alpha A_i^{t-1} \qquad (4)$$

$$r_i^t = r_i^0[1 - \exp(-\gamma(t-1))] \qquad (5)$$

Pseudo-code of the Standard Bat Algorithm
---

Set the Initial parameters:

Number of Population: P

Max number of iterations (Generations): T

Define pulse frequency $f_i$ for each bat at the position $x_i$

Define each bat's velocity $v_i$, Loudness $A_i$ and pulse rate $r_i$

Set the problem dimensions: $d$

Define the Objective Function $f(x)$, where $x = (x_1, x_2, .... x_d)^T$

Repeat

    For each bat in the population

        create new positions (solutions) through changing of frequency,

        velocities and solutions ((Equations (1)–(3))

        select a random number: rand

        If ( rand $> r_i$ )

            choosing a new solution among the best solutions

            choosing a local solution near the selected best solutions

        end If

        else If

        create a new solution by flying randomly

        If (rand $< A_i$ & $f(x_i) < f(x_*)$)

            The new solution regarded as the best solution

            Increase the $r_i$ and reduce $A_i$

        end If

    end for

        Reorder the bats and obtain the current best solution x*

until (t ≠ maximum number of generations)

## 4. Bat Algorithm Limitations

Bat Algorithm was used successfully for continuous constrained optimization problems with a significant increase in the convergence rate to the global optimal solution, especially, if it's parameters are well-tuned. BA approaches the most promising solution areas, increasing the chance of a balance between the exploration and exploitation phase. However, it has some weaknesses that may negatively affect its performance::

1) One of the important factors through which the efficacy of most swarm intelligence algorithms is determined is its ability to produce a mutation. Besides, the diversity of the population achieves the continued ability to develop in the algorithm, but Bat Algorithm lacks the above features and thus cannot jump out from the local optima area, where the diversity decreases whenever the problem dimensions increase [26].

2) The value of loudness and pulse rate is changing by a linear equation, thus the exploration phase in Bat Algorithm is applied based on an uncomplicated random distorted scheme, which leads to being stuck in the local optima. Without modifications and updates to these two equations, it could be difficult to use the Bat Algorithm for different applications [26].

3) In the bat motion step, all the search agents move toward the best solution (position with best fitness value) and thus leads to falling into local optima. In addition to lacking the parameters that increase its exploration ability and prevent it from falling into this problem [27].

4) Another determinant is that in the standard Bat Algorithm, the movement to a new location (a new solution) is not performed unless it is better than the current global solution and this leads to limiting the movement of search agents [27].

5) BA does not support the optimization problems with discrete values and needs to improve its implementation, as well as requiring different parameter tuning [28].

## 5. Applications of Bat Algorithm

Bat Algorithm has been successfully applied to a wide variety of academic and industrial domains and the most important application's classes are shown in (Table 1): - it was used in the design of the welded beam. The results indicated that the proposed new algorithm performs efficiently for multi-objective optimization.

**Table 1: Bat Algorithm Applications**

| Domains | Applications |
| --- | --- |
| Classification | In (2012), a new classification model (Bat-FLANN) [29] has been presented by Mishra, Shaw, Mishra which is used to modify the weights of the FLANN classifier. For performance assessment, the suggested model compared with FLANN and PSO-FLANN and showed it's outperforming the mentioned classifiers because of the super echolocation ability of the Bat Algorithm. |
| | S Lu, X Qiu, J Shi, N Li, ZH Lu, P Chen, et al (2017) [30] proposed a classification system named (PBD) to diagnose brain health. The system has been used to efficiently classify healthy brains from diseased ones. A set of magnetic resonance (MR) images has been considered. In the beginning, the discrete wavelet transform (DWT) was implemented twice on several medical images dataset to obtain several sub-band coefficients, then the entropy has been used to minimize the number of features of resulted sub-bands coefficients and finally, the standard Bat Algorithm has been used to optimize the extreme machine learning for classification purpose, and because k-fold cross-validation is more suitable for the small data set, it has been applied to the sample images for performance evaluation. |
| Multiobjective | Yang (2012) [31] has introduced a new method of modified Bat Algorithm named multi-objective Bat Algorithm (MOBA) for solving multi-objective optimization problems. Firstly, the MOBA was used to solve a subset of test functions and then applied on the multi-objective design optimization problem in structural engineerings, |

In (2012), Two types of an optimization problem: single and multi-objective brushless DC wheel motor problems have been optimized using the new proposed Bat Algorithm by Bora, Coelho, and Lebensztejn [32]. Both with 5 design parameters but 6 constraints in the single-objective problem and 2 objectives, 5 constraints in the multi-objective problem. The experimental results of comparing BA for single-objective problems with other algorithms indicated BA outperforming in most cases, and because BA combines the good features of two well-known algorithms: PSO and SA, it proved the best performance in the two analyzed multi-objective problems with best the results of Pareto Front.

Ramesh, Mohan, and Reddy (2013) [33] applied Bat Algorithm to solve the energy distribution problem economically to obtain an optimal integration for all generating units which consequently reduces the total cost of fuel, taking into account operational parameters and load burdens. For validation, CEED problem with different generating units has been tested, also a comparison with other algorithms has been conducted with significant results of BA which due to its computational accuracy and efficiency, as well as its high convergence rate and solutions performance.

| | |
|---|---|
| Load Balancing | Sharma, Luhanch, and Abdhullah (2016) [34] proposed a load balancing algorithm on the Bat Algorithm, which was applied to collect the goals of a load balancer. A toolbox implemented the proposed algorithm in Matlab named "Parallel Processing toolbox for validation purposes, the Round Robin technique, and Fuzzy GSO have been used for comparison to the suggested algorithm. As the final result was discussed in response time. While balancing the load between virtual machines, the response time was affected by job migration to another virtual machine. The Bat Algorithm in various virtual machines helped in balancing the load effectively. |
| Test Estimation Effort | Srivastava, Bidwai, Khan, Rathore, Rohit, and Yang (2014) [35] implemented BA to optimize the weights of parameters and then using them to predict the test effort of any project. For validation of the suggested model, three case studies (UCP, TPA, and industrial data ) have been tested, also compared with other algorithms (PSO and cuckoo search). The presented model estimated the test effort nearer to the real effort and more precisely than other algorithms. |
| Cloud Computing | Raghavan, Marimuthu, Sarwesh, Chandrasekaran (2015) [36] proposed a Bat Algorithm for scheduling workflow applications in cloud computing. The aim of such applications is assigning jobs to the available resources in an optimal manner with minimum cost achieved for the workflow. To select a minimum cost, BA has been implemented, while BBA with simple modifications has been used to select features. From the given results, BA provided less cost than the BRS algorithm. |
| Feature Selection | To select the best set of features, Nakamura, Pereira, Costa, Rodrigues, and Papa (2012) [37] combined the strong exploration ability of BBA with the fast convergence rate of Optimum-Path Forest classifier. Five general datasets have been utilized and they compared BA with PSO, FFA, and GSA. As a result, they found that the new approach |
|  | could perform in a good way for swarm-based techniques and it was a robustness algorithm. Also, it had good generalization capability. |
| Attribute Reduction | Recently, attribute reduction methods have been widely used in several fields like data mining, image processing, pattern recognition, etc. In (2013), Taha and Tang [38] proposed a new method called BAAR, where the Bat Algorithm has been used to select the more promising subset of attributes from a group of possible attributes. The efficiency of BAAR due to using a large number of parameters compared with other methods. They verified the BAAR on 13 public datasets; the experimental outcomes show an equal or better performance of BAAR than other feature selection methods, which showed that BAAR was a promising attribute reduction technique. |
| Image Processing | Bat Algorithm is also used in image and video processing. In 2012, S. Akhtar, A.R. Ahmad, E. M. Abdel-Rahman [39] used the Bat Algorithm to address the problem of tracking human body movement taken from a recorded video section. As it's known, Bat Algorithm takes the advantages of PSO and the Harmony Search, it combines the best features of PSO in exploration ability and the deep local search of the Bat Algorithm. As well as, the Bat algorithm, has a low computation cost. Appropriate changes on the standard Bat Algorithm parameters are have been made to fit with the requirement of the mentioned problem. To evaluate the algorithm performance against other optimization algorithms, fidelity criteria analyses have been used with noticeable outperforming of the Bat algorithm. |
| Fuel and Energy | The need to save energy and find the optimal methods to utilize its sources led to many research and studies on energy converting devices. In 2011, Tamiru L., and Fakhruldin H. [40] used the fuzzy system and Bat Algorithm to find the energy devastation distinctions in the gas turbine, Firstly, the fuzzy model is addressed and then modified Bat Algorithmis used to train the fuzzy model and to adjust its parameters. |
| Swarm Robotics | P Suárez, A Iglesias, A Gálvez (2019 ) claim that there are no previous studies in the field of applying the Bat Algorithm in swarm robotics, so they implemented the Bat Algorithm in swarm robotics through two stages: In the first stage, a prototype of real robotic has been designed and constructed; While in the second stage, a simulation framework of robotic has been developed. The simulation has been tested using the problem of finding a specific location inside 3-D places. The results of this study indicated that there are very similarities between the real robotic swarms and the implemented simulation. The contribution of this study has been to provide an optimal tool to highlight Bat Algorithms' limitations for different real-world applications within their simulations [41]. |

The purposes of Bat Algorithm applications and their performance to solve several optimization problems have been summarized in (Table 2).

**Table 2: The purposes of Bat Algorithm applications and their performance**

| No. | Publication Year | Purpose | Performance | Reference |
|---|---|---|---|---|
| 1. | 2012 | Produce a classification model based on the Bat Algorithm to optimize the weights of (FLANN) classifier. | BA Outperformed PSO | [29] |
| 2. | 2017 | To classify pathological brain MR images from healthy ones using 2D_DWT for feature extraction and optimized extreme learning machine with Bat Algorithm for classification. | BA_ELM outperformed four types of MR Brain Detection systems (DWT+PCA+SVM [42], WE+NBC [43], DWT + PCA + BPNN [43], DWT + PCA + RBFNN [43]) in terms of sensitivity and overall precision.. | [30] |
| 3. | 2012 | Solving the multiobjective optimization problem (Welded Beam Design). | A Pareto set functions (ZDT1, ZDT2, ZDT3, LZ4) was used to validate the efficiency of (MOBA) with remarkable convergence rate | [31] |
| 4. | 2012 | Solving a single and multiobjective optimization problem (DC Wheel Motor Problem). | For single-objective optimization problems, BA competitive against PSO and SA, and because it combines the features of PSO, SA, and NSGA-II, it also showed good performance for multiobjective problems. | [32] |
| 5. | 2013 | Design optimization model to eliminate the harmful emissions of fossil-fuelled power systems as well as solving economic problems of these systems. | For the CEED problem, Bat Algorithm outperformed RGA, SGA, Hybrid GA, and ABC | [33] |
| 6. | 2016 | Utilize Bat Algorithm to obtain the best load balancer server suitable for a particular task. | Significant decrease in response time of the proposed algorithm against the Round Robin and Fuzzy GSO algorithm. | [34] |
| 7. | 2014 | To optimise the weights of UCP and TPA parameters which can be | The experimental results indicated the proposed approach was more accurate and closer to the real effort than | [35] |
| | | later used to examine the effort estimation of similar jobs. | standard PSO and Cuckoo Search algorithms. | |
| 8. | 2015 | Bat Algorithm was used to choose the optimal tasks and resources which achieve the lowest total cost and putting them in the workflow scheduling in cloud computing. | To prove the Bat Algorithm's superiority, it compared with Best Resources Selection (BRS) algorithm with 50 % less cost over the (BRS) algorithm. | [36] |
| 9. | 2012 | Utilize the exploration ability of the Bat Algorithm and the high convergence rate of the Optimum Path Forest (OPF) classifier to get an optimal set of features. | On 5 general datasets, the proposed approach has been validated against 5 well-known binary meta-heuristic algorithms (PSO, FFA, HS, and GSA) and showed a top spot in three of the datasets while ranking second in the remaining datasets. | [37] |
| 10. | 2015 | Bat Algorithm has been used to optimize the real-time Fuzzy Inference System of DC motor. | PSO and Genetic Algorithm ANFIS optimizers have been used for validation. The proposed controller showed a considerably superior in terms of less fitness value, less computational duration, and high-quality performance in the majority working conditions. | [44] |
| 11. | 2015 | To design a Bat Algorithm with optimal parameter setting suitable for different numerical problems that have not any constraint. | The suggested Bat Algorithm with optimal parameter setting has been compared with 4 well-known modified Bat Algorithms using 28 CEC 2013 benchmark functions. The statistical measurements (ranking and P-value) showed the outperforming of the suggested parameter setting. | [45] |
| 12. | 2013 | Using Bat Algorithm to select a minimum set of features that is more feasible for image processing, data mining or machine learning. | Different datasets and five feature reduction algorithms (SimRSAR, AntRSAR, GenRSAR, TSAR, SSAR) have been used for validation, in term of performance, BAAR achieved a high feature reduction rate compared to other algorithms. | [38] |
| 13. | 2012 | Utilize Bat Algorithm to track the full human body movement captured from a video sequence. The problem was a non-linear optimization problem with high dimensionality. | For validation, quantitative and qualitative analyses have been performed with other meta-heuristic algorithms (APF, PF, and PSO). The results demonstrated the efficiency of | [39] |

| | | | the Bat Algorithm against PF, APF and PSO. | |
|---|---|---|---|---|
| 14. | 2011 | To design a simulation model that uses the fuzzy system in Gas Turbine Generator and its actual applicated later for conditional monitoring. | The performance of the suggested fuzzy model applied for the training and test data was sıgnıfıcant satisfied as the using of the modified Bat Algorithm added a lot of effective contribution to improving the existing problem. | [40] |
| 15. | 2019 | Bat Algorithm has been applied for optimizing the routing process of swarm robotics. After designing and implementing the physical components of the robotic prototype, Bat Algorithm has been used to develop the simulation, which simulates the real bat's behaviour in the process of finding prey in nature. | The simulation was confirmed by using it for the problem of finding a specific location in an internal 3D environment and with a result similar to the real robotic. Thus, a clearer vision was given the capabilities and drawbacks of this algorithm to solve real-world problems. | [41] |

## 5. Performance Evaluation

There are several studies were conducted to assess the performance of the Bat Algorithm against other meta-heuristic algorithms, most important studies are: -

- Recently meta-heuristic algorithms have been used widely for mathematical optimization problems. To compare their performances, Arora and Singh (2013) [46] used Bat Algorithm, Cuckoo Search, and Firefly Algorithm for optimization purposes. The experimental outcomes indicated that Firefly algorithm outperformed both Cuckoo Search and Bat Algorithm in terms of reaching speed to the optimal solution, which is dependent on the generating of random numbers during the algorithm iterations, also it takes less computational time to find the optimal solution without being stuck in the local optima as in Bat Algorithm. As well as, BA unable to save the history of the best position of each bat, Therefore they change their positions without regard to the previous best position and thus may move away from the ideal solution.

- Yang (2014) [47] has compared the performance of the Bat Algorithm with the intermittent search strategy. Any metaheuristic algorithm efficiency depends on several global explorations and how to balance intensive economic exploitation. After simulations, found that the Bat Algorithm is better than the optimal intermittent search strategy and can achieve a better balance between exploitation and exploration with superior efficiency. Also, it proved that the Bat Algorithm requires much more function evaluations.

- Nazmus, Md. Wasi, Md Subbir, and Mohammad (2014) [48] made a comparative study between two nature-inspired algorithms: the Flower pollination algorithm and basic Bat algorithm, the first algorithm is inspired by the natural pollination operation of the flower, while the second is inspired by the natural behavior of bats in echolocation. The performance of these two algorithms has been tested on both unimodal and multimodal benchmark functions. Also, these test functions are low and high dimensional continuous functions. The final results presented that the Flower pollination algorithm outperformed the Bat Algorithm in the majority of the functions, this is due to the weak exploration ability of the Bat Algorithm and the possibility of convergence to the global optimal solution is almost impossible.

- To choose the most proper meta-heuristic algorithm for the optimization problem, the performance assessment of each one should be considered because most of them are novel and have relatively unknown performance. So, Alsariera, Alamri, Nasser, Majid, and Zamli (2014) [49] made a comparative survey between the Bacterial Foraging Optimization algorithm (BFO) and the standard BA. Several benchmark functions have been implemented to assess the performance of each algorithm within the same parameter setting and programming language. The results displayed that the BFO algorithm is more accurate in reaching the optimal solution compared to the other algorithms, while BA has a higher convergence rate.

- In 2016, Rajesh S., Deeptimanta O. and Satyabrata D. [50] made a study to compare five types of swarm intelligence algorithms; Particle Swarm Optimization (PSO), Bat Algorithm(BA), Cuckoo Search (CS), Bee Colony Optimization (BCO), and Firefly Algorithm (FA). In this paper, the features of each algorithm and its ability to solve various optimization problems have been explained.

## 7. Hybridizations and Modifications of Bat Algorithm

In this section, we concentrate on presenting all the recently published developments of BA; including the following subsections:

- BA Hybridizations: with K-Medoids, Back-propagation neural network, Harmony Search Algorithm, Differential Evaluation Strategies, Enhanced Particle Swarm Optimization, Cuckoo Search Algorithm.

- BA Modifications: including MBA, EBA, BAM, UCAV-BAM, and Nonlinear optimization.

### 7.1 Hybridizations of Bat Algorithm

To exceed the limitations of BA, it has been merged with other well-known algorithms; which is reported in the following subsections:

### 7.1.1. K-Medoids

Sood and Bansal (2013) [51] used BA to optimize the representative object in the first step of the K-medoids clustering algorithm. Since classifying the initial representative object selection is the main topic in clustering, the K-medoids algorithm is hybrid with Bat Algorithm. With the help of BA, the initial representation object is more accessible. As a result, they presented the difference between the K-Medoid Clustering technique with Bat Algorithm and K-Medoid itself. These two algorithms gave an optimal solution in a large dataset; this hybrid is very useful in swarm intelligence for optimization and data mining, it is a clear way to present the object in the data sets.

### 7.1.2. Back-propagation neural network

Syed M.Gillani (2016) [52] suggested three types of improvements on the standard Bat Algorithm: Gaussian Distribution based Bat Algorithm (BAGD), Simulated Annealing Bat algorithm (SABa), and Genetic Bat Algorithm (GBA). Then, to optimize the initial weights of various neural networks, these proposed algorithms hybridized with Backpropagation and Elman Recurrent Neural Network. The results proved the superiority of the hybridized algorithms concerning execution time and the precision of finding the optimal solution.

### 7.1.3. Harmony Search Algorithm
To increase the population diversity and getting rid of weak fitness solutions, Wang and Guo (2013) [53] proposed a new approach that takes the advantages of two meta-heuristic algorithms, the improved BA has been merged with HS, then evaluated using a group of unimodal and multimodal numerical optimization problems. The hybridization has been done by adding the mutation feature of HS into the positon updating process of BA which lead to obtaining more new feasible solutions and escaping from the local optima. It allowed the bats to learn from more various examples because with each iteration the bats' positions were updated and new harmonies were constructed to search in a more promising search space, which accelerates the global rate of convergence with preserving the basic BA's strong robustness. Experimentally found that combining Harmony Search with Bat Algorithm takes better advantage of the information found in the previous solutions. The comparison results to the other swarm-based algorithms showed that the proposed approach was a better performance than both BA and HS.

### 7.1.4. Differential Evaluation Strategies
For constrained problems, Meng X, Gao X, and Liu Y (2015) [54] suggested a new algorithm (BADE) that hybridize BA with DE. In this simulation, sounds from surrounding objects were completely exploited by bats, which already dependent on locating insects through reflected waves. In addition to making use of the mutation factor present in DE, as well as the use of the average velocity of the neighbour bats in an equation of bat velocity updating. To prove it's outperforming against other algorithms, a set of benchmark functions have been tested, also it was implemented for different single-objective engineering problems.

### 7.1.5. Enhanced Particle Swarm Optimization
Tawhid and Dsouza (2018) [55] combined improved PSO and a binary BA in a novel algorithm named (HBBEPSO). The proposed algorithm took advantage of the Bat Algorithm's ability for exploration by echolocation to select the important features and the improved PSO algorithm ability in fast convergence capability to the best solution in the search space. For validation, ın addition to comparing it to the standard BA, HBBEPSO was compared to other optimizers. Different statistical measurements have been used to evaluate the performance of the algorithm and showed its efficient optimizing ability to select an optimal feature group.

### 7.1.6. Cuckoo Search Algorithm
Cuckoo Search (CS) has its limitations and drawbacks, especially in the exploitation phase where it leads to low convergence to the global optima and trapped into the local optima as well as, Cuckoo Search algorithm depends on two key parameters: ($p_a$) and ($\lambda$), which are used to increase the convergence probability, but the value of these factors dramatically rises the number of loops and could be incapable to find the best solution and could not reach the solution in case large iterations. Therefore, M Shehab, A T Khader, M Laouchedi, and O Alomari in 2019 [56] hybridized it with the standard Bat Algorithm and used the new algorithm (CSBA) to solve optimization problems. The suggested algorithm works for numerical problems by determining the population of the reception nests in the standard Cuckoo Search and find the optimal solution using BA. The efficiency of this algorithm has been examined by a group of unimodal and multimodal benchmark functions. The results of this study indicate that the hybridized algorithm is outperformed the standard Cuckoo Search in the majority of functions, especially for local search.

## 7.2. Modifications of Bat Algorithm

In the following subsection, all the modifications that have been conducted of Bat Algorithm are listed:

### 7.2.1. MBA

Fister, Rauter, Yang, Ljubic, Fister Jr (2014) [57] proposed an MBA algorithm as an expert planner for sport coaching sessions; the plan contains training period of the same athlete includes base training sessions. They modified the original Bat Algorithm to solve the planning for sports exercise. They applied the MBA algorithm on a ten base training course for a specific athlete to plan the cycle training; the results showed the plan which predicted was comply high standard of cycle coach.

### 7.2.2. EBA

To overcome the flaws in BA, Yilmaz and Kucuksille (2014) [58] suggested a new algorithm called (EBA) that improves the exploration ability by making the value of the loudness $A$ and the pulse rate $r$ equals to the number of problem dimensions. The experimental results on 20 benchmark functions with different dimensions proved the efficiency of EBA against BA.

### 7.2.3. BAM

Zhang and Wang (2012) [59] have introduced a new modified Bat Algorithm to deal with the image recognition problem called Bat Algorithm with mutation (BAM). The realization procedure for the modified algorithm BAM was presented. The new approach increased the convergence speed to the optimal solution while preserving the strength and efficiency of the original algorithm. BAM has been compared with BA and other meta-heuristic optimization algorithms, such as DE and SGA. An effective way has been used in image processing, with the mutation for image matching Bat Algorithm is simple, and it is reducing the number of candidate positions of matching which leads to reducing the time of complexity method. The result of the experiment proved that BAM was more applicable and sufficient for image matching than the other models.

### 7.2.4. UCAV-BAM

Wang, et. Al (2012) [60,61] have proposed a new Bat Algorithm with a mutation to solve the route planning problems of unmanned warship aerial vehicles [62]. The standard Bat Algorithm has been used for solving the UCAV path planning problem before they applied a modification to the original BA to switch between bats in generating new solutions step. To avoid costing minimum fuel and the threat area, the UCAV can fly safely from the source to the destination by joining the selected nodes of the coordinates. Bat Algorithm mutation BAM is maintaining the power of BA and can speed up the global convergence speed. The achievements of the BAM were compared first with BA and then with other algorithms, as an outcome proved that the new algorithm BAM was more successful and promise in UCAV path planning than other methods.

### 7.2.5. Nonlinear Optimization

Kielkowics and Grela (2016) [63] have proposed two modifications to the Bat Algorithm, introduced a different scheme of acceptance of the newly found solutions also the velocity equation is modified. They modified the acceptance scheme to reduce the probability of acceptance of worse solutions, and they changed the velocity update equation by archive component and introducing the cognitive coefficient. Last step they formed the additional memory by storing the best solutions found during the optimization process. They compared the modified Bat Algorithm with the standard algorithm, and they tested the modifications in a few simulation experiments. As the suggested modification has been done on the linear velocity equation, thus it didn't affect the computational complexity or the fitness function evaluation of the algorithm.

## 8. Optimization problems solved by Bat Algorithm

In this section, we classify all studies conducted on the Bat Algorithm based on the type of optimization problem that was addressed by this algorithm, (Table 3).

**Table 3: Summary of Continuous and Discrete problems addressed by Bat Algorithm**

| Optimization | Problem | Performance | Publication Year & Ref. |
|---|---|---|---|
| Continuous Optimization | Group of test functions and Welded beam design. | Efficient for multi-objective optimization problems. | 2011, [64] |
| | • Pressure vessel design. <br> • Welded beam design. <br> • A tension-compression spring design. | Outperformed other optimization algorithms (e.g., PSO, GA, EA, and SA). | 2013, [65] |
| | Three well-known benchmark test functions, no real application. | Less computational time and more accuracy. | 2012, [66] |
| | 14 standard benchmark test functions, no real application. | Better than PSO, standard BA, ACO, and others in the majority of unimodal and multimodal benchmark functions. | 2013, [53] |
| | Track design for a driverless warship. | Convergence rate improvement while preserving the features of the original algorithm. | 2012, [60] |
| | 20 Classical CEC2005 benchmark functions. | Outperformed standard Bat Algorithm, Self-adaptive Differential Evolution, and Improved Bat Algorithm with significant results. | 2017, [67] |
| | 10 benchmark functions. | For different dimensions implementation, IBA was better than BA in 25 of 26 test functions. | 2013, [68] |
| | Unimodal multimodal benchmark and CEC 2014 test functions. | Better than standard Bat algorithm in most functions, but slower. | 2018, [69] |
| | 10 benchmark test functions. | NABA better than the standard Bat Algorithm for high dimensional optimization problems | 2014, [70] |
| | 18 unimodal, multimodal, and continuous benchmark test functions. | More efficiency and less computational time than DE, saDE, and others | 2016, [71] |
| Discrete Optimization | Reducing the cost of the early and late penalty of scheduling problem of a multi-step production line. | Better performance after using the proper statistical tools for parameter adjusting. | 2012, [72] |
| | Reducing the makespan in the permutation flow shop problem. | More convergence speed and population diversity but slower. | 2013, [73] |
| | Best sports training session plan for a specific athlete. | The proposed training plan is highly compatible with the standards of international coaches. | 2015, [57] |
| | 22 standard benchmark test functions, as well as photonic crystal waveguide problem. | BBA outperformed other binary heuristic algorithms in terms of fast convergence and avoiding local optima, also proved its efficiency in real application of optical engineering problem. | 2013, [74] |

|  | The Traveling Salesman Problem. | IBA better than other meta-heuristic optimization algorithms for symmetric and asymmetric Travelling Salesman Problem. | 2016, [75] |
|---|---|---|---|
|  | The Travelling Salesman problem, as well as 41 benchmark test functions. | DBA better than DPSO, GSA-ACS-PSOT, and IDCS. | 2016, [76] |
|  | knapsack problem | Better than GA, Improved PSO-R, ACO, and BCO in solving discrete knapsack problem | 2014, [77] |
|  | multi-attribute vehicle routing problem | Comparing with ESA, EA, and FA, 75% better performance and faster convergence. | 2019, [78] |
|  | permutation flow shop scheduling problem | Similar to PSO Based Memetic Algorithm, but higher convergence rate and less solution precision. | 2014, [79],[80] |
|  | Finding the community structure in networks where the modularity quality measure has been used as an objective function in Bat Algorithm. | Good performance against other community detection methods for small networks. | 2015, [27] |

## 9. Improved Bat Algorithm

During the past years, several improvements have been done on Bat Algorithm to improve its performance, this section includes all the studies found in the literature and gives a brief discussion of these improvements:

To optimize the multilevel image thresholds and to enhance the efficiency of Kapur's function, Alihodzic and Tuba (2014) [81] have proposed a new approach that merged BA with Kapur's method. The suggested method has been compared with the CS algorithm. Experiments conducted on a group of images showed that both algorithms have the same accuracy in performance, while BA was better in terms of computational time than CS and faster in the converging rate from the exhaustive algorithm.

B Bahmani-Firouzi, R Azizipanah-Abarghooee (2014) [82] introduced a new enhanced Bat Algorithm used to the advanced design of corrective planning with minimum transmission cost for the formulation which is cost-based. This study aimed to specify the optimal capacity size of Battery Energy in the operation management of Micro-Grid. The simulation of the above problem deduced several outcomes. First, for complex (Generalized Rastrigin, Generalized Grienwangk, and Generalized Ackley) benchmark functions and case study A, the experimental results indicated the outperforming of the Improved Bat Algorithm in several aspects including high convergence rate, less computational complexity, and best solutions. As well as, through the optimal size of MG and adding the Li-ion BES to the case study B, the total cost of the Micro-Grid could be significantly reduced. And finally, through comparing the outcomes of case study A, B, and C, it is obvious that when setting the optimal size of 150 kWh BES, the overall cost will be minimized to about 40 per cent in every day against the MG without BES, while setting the best size of 250 kW h and an initial charge to 250 kW h will minimize the cost 15 per cent every day.

The process of diagnosing a heart attack in people with heart diseases is done by detecting the changes occurring in the signal of an ECG, this process is preceded by an initial step to get rid of noise as well as extracting features that can be used in the diagnosis, in (2015) Kora and Kalva[83] proposed a modified approach of BA for feature extracting of heartbeats, then the optimized features have been implemented as an input of NN classification system. The obtained results have been compared with other heart attack detection algorithms with significant improvement of the classifier through using the best features deduced of IBA.

As the standard Bat Algorithm has exploitation and exploration ability weaknesses especially when it is utilized to solve multi-modal optimization problems, X Cai, X Gao, Y Xue (2016) [84] proposed an improved Bat Algorithm where a strong local search strategy based on the optimal forage method is designed to improve the exploitation phase. Moreover, to avoid local optima and enhancing global search capability, a new disturbance method has been used which increases the usage rate of information among Bats. The efficiency of the presented algorithm has been tested through making a comparison with several algorithms (including standard Bat, Levay Bat Algorithm, Particle Swarm Optimization, and Cuckoo Search). Also, 28 CEC2013 benchmark functions have been used for validation.

In 2019. E Osaba, XS Yang, I Fister Jr, J Del Ser [78] have presented an improved Bat Algorithm named (DalBat). In this development, Hamming Distance (HC) has been used to measure the path between the bats. Also, another approach has been adopted to the proposed algorithm by depending on two structures of the neighbourhood Bats, which can be found depending on the optimal search agent in the population. The authors also claim that it was used for the first time to solve a drug distribution problem in the real world in addition to the disposal of medical waste.

Also, In (2019) [85] WC Hong, MW Li, J Geng, Y Zhang presented a new model to estimate the movement of the floating platform. After the simulation of the motion of a floating platform has been implemented using a vector regression model, an improved Bat Algorithm has been suggested to optimize the operators of the model. And finally, to decompose the motion time series, the signal empirical mode decomposition technique (EMD) has been used. The contribution of this study has been to confirm the activity and reliability of the suggested model in FPM estimation.

Bat's low convergence ability and the possibility of falling into the local optimum, as mentioned in many of the previous research done on Bat Algorithm, led the researchers (M.R.Ramli, Z.Abal Abas, M.I.Desa, Z. Zainal Abidin, and M.B.Alazzam) in (2019) [86] to improve its exploitation capability through using a dynamic number of dimension size rather than a static number. This operation helped to select a more suitable random value for the next iterations, also the selected value ignored representing useless dimension region. To validate the accuracy of the modified algorithm, ten benchmark functions have been tested and the results indicated its better optimizing of the solution.

## 10. New Bat Algorithm

This section presents all the new adapted Bat Algorithms:

### 10.1. Directed Artificial Bat Algorithm

Rekaby (2013) [87] has introduced a new algorithm inspired by the bat's echolocation action named (DABA). For prey catching, the algorithm uses the behaviour of simulating the bat echo emitting and the refection capturing for having a figure of the entities around. The new algorithm has the same technique to deal with optimization problems, and it has the same concept of real bats for hunting. They compare DABA with an artificial bee colony, and the results suggested that DABA performed better than the ABC algorithm with an equal number of iterations and population size with per cent ranged 5 to %10 of fitness value improvements.

### 10.2. Complex-Valued Bat Algorithm

Li and Zhou (2014) [88] have introduced a new Bat Algorithm based on complex-valued encoding (CBA). The new approach expanded the dimensions for denoting as well as the diversity of the population has been increased, it was also improved the optimization efficiency of the algorithm. They tested the new proposed algorithm on the convergence benchmark test function. After the simulation, the results indicated the outperforming of the proposed algorithm. When comparing the CBA with BA and PSO, it was better in terms of convergence rate, the accuracy of optimization, and the power of CBA against BA and PSO.

### 10.3. Principal Component Analyses combined with Bat Algorithm

In the standard Bat Algorithm, the high ratio of search agents (Bats) that are strongly correlated negatively affects the efficiency of the algorithm and thus reduces the population diversity and consequently, it causes the prevention of global search. Wherefore, Z Cui, F Li, W Zhang (2018) [89] proposed a new algorithm in which Principal Component Analyses are merged with Bat Algorithm(PCA-BA). PCA is used to analyze all the search agents. To separate highly correlated individuals from the low correlated individuals in the population appropriate threshold are used and only individuals with high correlation were taken into account by PCA. CEC2008 benchmark functions showed that the new algorithm outperformed the standard one, also it was compared with other well-known evolutionary optimization algorithms with satisfactory results.

### 10.4. A new Bat Algorithm for Numerical Optimization

The convergence rate of the optimal solution in the exploration phase of the Bat Algorithm declines whenever the parameters are increases, so Y Wang, P Wang, J Zhang, Z Cui, X Cai, W Zhang in (2019) [90] proposed a new algorithm called (MixBAT) which used eight selection strategy depending on the value of the fitness function. There were significant outperforming results of MixBat on other evolutionary algorithms validated through CEC2013 benchmark test functions. Although, this method achieves the global search and avoids the local optima effectively. However, sometimes other strategies could be cancelled in the initial iterations and trapped in the local optima.

### 10.5. Solving continuous optimization problems by directional Bat Algorithm

To improve the standard Bat Algorithm performance and to overcome its limitations, a new directional algorithm named (dBA) has been presented by (Asma CHAKRI, Rabia KHELIF, Mohamed BENOUARET, Xin-She YANG)(2016) [67] which four changes have been made on the original Bat Algorithm. To direct the search mechanism towards the more useful areas, they used the direction of the best position (which determined by objective fitness). To show the outperforming of the suggested algorithm (dBA), three groups of experiments with different benchmark functions have been used.

## 11. Conclusion and Future Work

This research aimed to highlight on Bat Algorithm and its advancement and applications. In the beginning, the biological nature of Bat and its hunting behaviour through issuing echolocation signals were explained in detail. Furthermore, the standard algorithm proposed by Yang (BA) was presented with its original pseudo-code and equations. After that, all the limitations reported on the algorithm in previous studies were detailed. As well, all applications and problems that have been addressed using the algorithm to solve various optimization problems are explained. Also, for more convergence analyses, the performance of (BA) was assessed against other meta-heuristic nature-inspired algorithms, where the study included a comparison with the five most important studies conducted in this field. Additionally, the survey contained all the hybridizations and modifications studies of (BA). Finally, all improved and new Bat Algorithms were summarized. BA proved its ability in handling various and wide fields optimization whether discrete or continuous problems. However, as with other meta-heuristic optimization problems, BA has its drawbacks with high dimensional problems and is often trapped to the local optima. As well as lack of population diversity and mutation ability. The study contributes to the researchers' understanding of the algorithm completely and opening new perspectives in its field of development. The success of this algorithm and other metaheuristic algorithms will greatly increase the number of research studies in this field for their development. The problem of falling into the local optima, as well as achieving a balance between exploration and exploitation phases are considered important issues that must be focused on in the future. Also, tracking the local optimum obtained in previous iterations or the use of modern clustering techniques may be a new area of research.

# Compliance with Ethical Standards

Conflict of interest: The authors declare that they have no conflict of interest.

# Funding

Funding information is not applicable / No funding was received.

# Bibliography

1. Beni G, Wang J (1993) Swarm intelligence in cellular robotic systems. In: Robots and biological systems: towards new bionics? Springer, pp 703-712
2. Hassan BA, Rashid TA (2020) Operational framework for recent advances in backtracking search optimisation algorithm: a systematic review and performance evaluation. Applied Mathematics and Computation 370:124919
3. Yang X-S (2010) Nature-inspired metaheuristic algorithms. Luniver press,
4. Rajakumar R, Dhavachelvan P, Vengattaraman T A survey on nature-inspired meta-heuristic algorithms with its domain specifications. In: 2016 International Conference on Communication and Electronics Systems (ICCES), 2016. IEEE, pp 1-6
5. Ahmed AM, Rashid TA, Saeed SAM (2020) Cat Swarm Optimization Algorithm: A Survey and Performance Evaluation. Computational Intelligence and Neuroscience 2020
6. Mitchell M (1998) An introduction to genetic algorithms. MIT press,
7. Kirkpatrick S, Gelatt CD, Vecchi MP (1983) Optimization by simulated annealing. science 220 (4598):671-680
8. Kennedy J, Eberhart R Particle swarm optimization. In: Proceedings of ICNN'95-International Conference on Neural Networks, 1995. IEEE, pp 1942-1948
9. Abdullah JM, Ahmed T (2019) Fitness dependent optimizer: inspired by the bee swarming reproductive process. IEEE Access 7:43473-43486
10. Dorigo M, Di Caro G Ant colony optimization: a new meta-heuristic. In: Proceedings of the 1999 congress on evolutionary computation-CEC99 (Cat. No. 99TH8406), 1999. IEEE, pp 1470-1477
11. Storn R, Price K (1997) Differential evolution–a simple and efficient heuristic for global optimization over continuous spaces. Journal of global optimization 11 (4):341-359
12. Karaboga D (2005) An idea based on honey bee swarm for numerical optimization. Technical report-tr06, Erciyes university, engineering faculty, computer …,
13. Yang X-S Firefly algorithms for multimodal optimization. In: International symposium on stochastic algorithms, 2009. Springer, pp 169-178
14. Yang X-S, Deb S Cuckoo search via Lévy flights. In: 2009 World congress on nature & biologically inspired computing (NaBIC), 2009. IEEE, pp 210-214
15. Yang X-S (2010) A new metaheuristic bat-inspired algorithm. In: Nature inspired cooperative strategies for optimization (NICSO 2010). Springer, pp 65-74
16. Yu B, Cui Z, Zhang G (2013) Artificial plant optimization algorithm with correlation branches. Journal of Bioinformatics and Intelligent Control 2 (2):146-155
17. Mirjalili S (2016) Dragonfly algorithm: a new meta-heuristic optimization technique for solving single-objective, discrete, and multi-objective problems. Neural Computing and Applications 27 (4):1053-1073

18. Mirjalili S, Lewis A (2016) The whale optimization algorithm. Advances in engineering software 95:51-67
19. Mirjalili S, Gandomi AH, Mirjalili SZ, Saremi S, Faris H, Mirjalili SM (2017) Salp Swarm Algorithm: A bio-inspired optimizer for engineering design problems. Advances in Engineering Software 114:163-191
20. Cortés-Toro EM, Crawford B, Gómez-Pulido JA, Soto R, Lanza-Gutiérrez JM (2018) A new metaheuristic inspired by the vapour-liquid equilibrium for continuous optimization. Applied Sciences 8 (11):2080
21. Shamsaldin AS, Rashid TA, Al-Rashid Agha RA, Al-Salihi NK, Mohammadi M (2019) Donkey and smuggler optimization algorithm: A collaborative working approach to path finding. Journal of Computational Design and Engineering 6 (4):562-583
22. Yang B, Wang J, Zhang X, Yu T, Yao W, Shu H, Zeng F, Sun L (2020) Comprehensive overview of meta-heuristic algorithm applications on PV cell parameter identification. Energy Conversion and Management 208:112595
23. Rahman CM, Rashid TA (2020) A new evolutionary algorithm: Learner performance based behavior algorithm. Egyptian Informatics Journal
24. Metzner W (1991) Echolocation behaviour in bats. Science Progress (1933-):453-465
25. Mohammed HM, Umar SU, Rashid TA (2019) A systematic and meta-analysis survey of whale optimization algorithm. Computational intelligence and neuroscience 2019
26. Jun L, Liheng L, Xianyi W (2015) A double-subpopulation variant of the bat algorithm. Applied Mathematics and Computation 263:361-377
27. Hassan EA, Hafez AI, Hassanien AE, Fahmy AA A discrete bat algorithm for the community detection problem. In: International Conference on Hybrid Artificial Intelligence Systems, 2015. Springer, pp 188-199
28. Ezugwu AE, Adeleke OJ, Akinyelu AA, Viriri S (2020) A conceptual comparison of several metaheuristic algorithms on continuous optimisation problems. Neural Computing and Applications 32 (10):6207-6251
29. Mishra S, Shaw K, Mishra D (2012) A new meta-heuristic bat inspired classification approach for microarray data. Procedia Technology 4:802-806
30. Lu S, Qiu X, Shi J, Li N, Lu Z-H, Chen P, Yang M-M, Liu F-Y, Jia W-J, Zhang Y (2017) A pathological brain detection system based on extreme learning machine optimized by bat algorithm. CNS & Neurological Disorders-Drug Targets (Formerly Current Drug Targets-CNS & Neurological Disorders) 16 (1):23-29
31. Yang X-S (2012) Bat algorithm for multi-objective optimisation. arXiv preprint arXiv:12036571
32. Bora TC, Coelho LdS, Lebensztajn L (2012) Bat-inspired optimization approach for the brushless DC wheel motor problem. IEEE Transactions on magnetics 48 (2):947-950
33. Ramesh B, Mohan VCJ, Reddy VV (2013) Application of bat algorithm for combined economic load and emission dispatch. Int J of Electricl Engineering and Telecommunications 2 (1):1-9
34. Sharma S, Luhach AK, Abdhullah SS (2016) An optimal load balancing technique for cloud computing environment using bat algorithm. Indian Journal of Science and Technology 9 (28):1-4
35. Srivastava PR, Bidwai A, Khan A, Rathore K, Sharma R, Yang XS (2014) An empirical study of test effort estimation based on bat algorithm. International Journal of Bio-Inspired Computation 6 (1):57-70
36. Raghavan S, Sarwesh P, Marimuthu C, Chandrasekaran K Bat algorithm for scheduling workflow applications in cloud. In: 2015 International Conference on Electronic Design, Computer Networks & Automated Verification (EDCAV), 2015. IEEE, pp 139-144

37. Nakamura RY, Pereira LA, Costa KA, Rodrigues D, Papa JP, Yang X-S BBA: a binary bat algorithm for feature selection. In: 2012 25th SIBGRAPI conference on graphics, Patterns and Images, 2012. IEEE, pp 291-297
38. Taha AM, Tang AY (2013) Bat algorithm for rough set attribute reduction. Journal of theoretical and applied information technology 51 (1):1-8
39. Akhtar S, Ahmad A, Abdel-Rahman EM A metaheuristic bat-inspired algorithm for full body human pose estimation. In: 2012 Ninth Conference on Computer and Robot Vision, 2012. IEEE, pp 369-375
40. Lemma TA, Hashim FBM Use of fuzzy systems and bat algorithm for exergy modeling in a gas turbine generator. In: 2011 IEEE Colloquium on Humanities, Science and Engineering, 2011. IEEE, pp 305-310
41. Suárez P, Iglesias A, Gálvez A (2019) Make robots be bats: specializing robotic swarms to the bat algorithm. Swarm and evolutionary computation 44:113-129
42. Zhang Y-D, Wu L (2012) An MR brain images classifier via principal component analysis and kernel support vector machine. Progress In Electromagnetics Research 130:369-388
43. Zhou X, Wang S, Xu W, Ji G, Phillips P, Sun P, Zhang Y Detection of pathological brain in MRI scanning based on wavelet-entropy and naive Bayes classifier. In: International conference on bioinformatics and biomedical engineering, 2015. Springer, pp 201-209
44. Premkumar K, Manikandan B (2015) Speed control of Brushless DC motor using bat algorithm optimized Adaptive Neuro-Fuzzy Inference System. Applied Soft Computing 32:403-419
45. Xue F, Cai Y, Cao Y, Cui Z, Li F (2015) Optimal parameter settings for bat algorithm. International Journal of Bio-Inspired Computation 7 (2):125-128
46. Arora S, Singh S A conceptual comparison of firefly algorithm, bat algorithm and cuckoo search. In: 2013 International Conference on Control, Computing, Communication and Materials (ICCCCM), 2013. IEEE, pp 1-4
47. Yang X-S, Deb S, Fong S (2014) Bat algorithm is better than intermittent search strategy. arXiv preprint arXiv:14085348
48. Sakib N, Kabir MWU, Subbir M, Alam S (2014) A comparative study of flower pollination algorithm and bat algorithm on continuous optimization problems. International Journal of Soft Computing and Engineering 4 (3):13-19
49. Alsariera YA, Alamri HS, Nasser AM, Majid MA, Zamli KZ Comparative performance analysis of bat algorithm and bacterial foraging optimization algorithm using standard benchmark functions. In: 2014 8th. Malaysian Software Engineering Conference (MySEC), 2014. IEEE, pp 295-300
50. Sahoo RK, Ojha D, Dash S (2016) NATURE INSPIRED METAHEURISTIC ALGORITHMS-A COMPARATIVE REVIEW.
51. Sood M, Bansal S (2013) K-medoids clustering technique using bat algorithm. International Journal of Applied Information Systems 5 (8):20-22
52. Rehman Gillani SMZ (2016) An improved bat algorithm with artificial neural networks for classification problems. Universiti Tun Hussein Onn Malaysia,
53. Wang G, Guo L (2013) A novel hybrid bat algorithm with harmony search for global numerical optimization. Journal of Applied Mathematics 2013
54. Meng X, Gao X, Liu Y (2015) A novel hybrid bat algorithm with differential evolution strategy for constrained optimization. International Journal of Hybrid Information Technology 8 (1):383-396
55. Tawhid MA, Dsouza KB (2018) Hybrid binary bat enhanced particle swarm optimization algorithm for solving feature selection problems. Applied Computing and Informatics

56. Shehab M, Khader AT, Laouchedi M, Alomari OA (2019) Hybridizing cuckoo search algorithm with bat algorithm for global numerical optimization. The Journal of Supercomputing 75 (5):2395-2422

57. Fister I, Rauter S, Yang X-S, Ljubič K, Fister Jr I (2015) Planning the sports training sessions with the bat algorithm. Neurocomputing 149:993-1002

58. Yılmaz S, Kucuksille EU, Cengiz Y (2014) Modified bat algorithm. Elektronika ir Elektrotechnika 20 (2):71-78

59. Zhang JW, Wang GG Image matching using a bat algorithm with mutation. In: Applied Mechanics and Materials, 2012. Trans Tech Publ, pp 88-93

60. Wang G, Guo L, Duan H, Liu L, Wang H (2012) A bat algorithm with mutation for UCAV path planning. The Scientific World Journal 2012

61. Ali ZA, Zhangang H, Zhengru D (2020) Path planning of multiple UAVs using MMACO and DE algorithm in dynamic environment. Measurement and Control:0020294020915727

62. Ali ZA, Zhangang H, Hang WB (2020) Cooperative Path Planning of Multiple UAVs by using Max–Min Ant Colony Optimization along with Cauchy Mutant Operator. Fluctuation and Noise Letters:2150002

63. Kiełkowicz K, Grela D (2016) Modified Bat algorithm for nonlinear optimization. International Journal of Computer Science and Network Security (IJCSNS):46-50

64. Yang X-S (2011) Bat algorithm for multi-objective optimisation. International Journal of Bio-Inspired Computation 3 (5):267-274

65. Gandomi AH, Yang X-S, Alavi AH, Talatahari S (2013) Bat algorithm for constrained optimization tasks. Neural Computing and Applications 22 (6):1239-1255

66. Tsai PW, Pan JS, Liao BY, Tsai MJ, Istanda V Bat algorithm inspired algorithm for solving numerical optimization problems. In: Applied mechanics and materials, 2012. Trans Tech Publ, pp 134-137

67. Chakri A, Khelif R, Benouaret M, Yang X-S (2017) New directional bat algorithm for continuous optimization problems. Expert Systems with Applications 69:159-175

68. Yilmaz S, Kucuksille EU (2013) Improved bat algorithm (IBA) on continuous optimization problems. Lecture Notes on Software Engineering 1 (3):279

69. Liu Q, Wu L, Xiao W, Wang F, Zhang L (2018) A novel hybrid bat algorithm for solving continuous optimization problems. Applied Soft Computing 73:67-82

70. Kabir MWU, Sakib N, Chowdhury SMR, Alam MS (2014) A novel adaptive bat algorithm to control explorations and exploitations for continuous optimization problems. International Journal of Computer Applications 94 (13)

71. Pravesjit S (2016) A hybrid bat algorithm with natural-inspired algorithms for continuous optimization problem. Artificial Life and Robotics 21 (1):112-119

72. Musikapun P, Pongcharoen P Solving multi-stage multi-machine multi-product scheduling problem using bat algorithm. In: 2nd international conference on management and artificial intelligence, 2012. IACSIT Press Singapore, pp 98-102

73. Xie J, Zhou Y, Tang Z Differential lévy-flights bat algorithm for minimization makespan in permutation flow shops. In: International Conference on Intelligent Computing, 2013. Springer, pp 179-188

74. Mirjalili S, Mirjalili SM, Yang X-S (2014) Binary bat algorithm. Neural Computing and Applications 25 (3-4):663-681

75. Osaba E, Yang X-S, Diaz F, Lopez-Garcia P, Carballedo R (2016) An improved discrete bat algorithm for symmetric and asymmetric traveling salesman problems. Engineering Applications of Artificial Intelligence 48:59-71

76. Saji Y, Riffi ME (2016) A novel discrete bat algorithm for solving the travelling salesman problem. Neural Computing and Applications 27 (7):1853-1866
77. Sabba S, Chikhi S (2014) A discrete binary version of bat algorithm for multidimensional knapsack problem. International Journal of Bio-Inspired Computation 6 (2):140-152
78. Osaba E, Yang X-S, Fister Jr I, Del Ser J, Lopez-Garcia P, Vazquez-Pardavila AJ (2019) A discrete and improved bat algorithm for solving a medical goods distribution problem with pharmacological waste collection. Swarm and evolutionary computation 44:273-286
79. Luo Q, Zhou Y, Xie J, Ma M, Li L (2014) Discrete bat algorithm for optimal problem of permutation flow shop scheduling. The Scientific World Journal 2014
80. Liu B, Wang L, Jin Y-H (2007) An effective PSO-based memetic algorithm for flow shop scheduling. IEEE Transactions on Systems, Man, and Cybernetics, Part B (Cybernetics) 37 (1):18-27
81. Alihodzic A, Tuba M (2013) Bat algorithm (BA) for image thresholding. Recent Researches in Telecommunications, Informatics, Electronics and Signal Processing:17-19
82. Bahmani-Firouzi B, Azizipanah-Abarghooee R (2014) Optimal sizing of battery energy storage for micro-grid operation management using a new improved bat algorithm. International Journal of Electrical Power & Energy Systems 56:42-54
83. Kora P, Kalva SR (2015) Improved Bat algorithm for the detection of myocardial infarction. SpringerPlus 4 (1):666
84. Cai X, Gao X-z, Xue Y (2016) Improved bat algorithm with optimal forage strategy and random disturbance strategy. International Journal of Bio-Inspired Computation 8 (4):205-214
85. Hong W-C, Li M-W, Geng J, Zhang Y (2019) Novel chaotic bat algorithm for forecasting complex motion of floating platforms. Applied Mathematical Modelling 72:425-443
86. Ramli M, Abas ZA, Desa M, Abidin ZZ, Alazzam M (2019) Enhanced convergence of Bat Algorithm based on dimensional and inertia weight factor. Journal of King Saud University-Computer and Information Sciences 31 (4):452-458
87. Rekaby A Directed Artificial Bat Algorithm (DABA)-A new bio-inspired algorithm. In: 2013 International Conference on Advances in Computing, Communications and Informatics (ICACCI), 2013. IEEE, pp 1241-1246
88. Li L, Zhou Y (2014) A novel complex-valued bat algorithm. Neural Computing and Applications 25 (6):1369-1381
89. Cui Z, Li F, Zhang W (2019) Bat algorithm with principal component analysis. International Journal of Machine Learning and Cybernetics 10 (3):603-622
90. Wang Y, Wang P, Zhang J, Cui Z, Cai X, Zhang W, Chen J (2019) A novel bat algorithm with multiple strategies coupling for numerical optimization. Mathematics 7 (2):135